\documentclass[runningheads]{llncs}
\usepackage[T1]{fontenc}
\usepackage{amsmath}
\usepackage{amssymb}
\usepackage{wrapfig}
\usepackage[table]{xcolor} % Required for row colors
\usepackage{multirow}
\usepackage{booktabs}
\usepackage{graphicx}  % for \scalebox
\usepackage{xcolor}
\usepackage{colortbl}  % for \rowcolor
\usepackage{pifont}    % for checkmarks
\usepackage{hyperref}
\hypersetup{
    colorlinks=true,
    urlcolor=blue
}
\definecolor{oursgray}{gray}{0.9}

\newcommand{\cmark}{\ding{51}}
\newcommand{\xmark}{\ding{55}}

\usepackage{graphicx,verbatim}
\begin{document}
\title{TriFusion-SR: Joint Tri-Modal Medical Image Fusion and SR}
\begin{comment}  %% Removed for anonymized MICCAI submission
\author{First Author\inst{1}\orcidID{0000-1111-2222-3333} \and
Second Author\inst{2,3}\orcidID{1111-2222-3333-4444} \and
Third Author\inst{3}\orcidID{2222--3333-4444-5555}}
%
\authorrunning{F. Author et al.}
\institute{Princeton University, Princeton NJ 08544, USA \and
Springer Heidelberg, Tiergartenstr. 17, 69121 Heidelberg, Germany
\email{lncs@springer.com}\\
\url{http://www.springer.com/gp/computer-science/lncs} \and
ABC Institute, Rupert-Karls-University Heidelberg, Heidelberg, Germany\\
\email{\{abc,lncs\}@uni-heidelberg.de}}

\end{comment}
\author{
Fayaz Ali Dharejo\inst{1} \and
Sharif S. M. A.\inst{2} \and
Aiman Khalil\inst{3} \and\\
Nachiket Chaudhary\inst{1} \and
Rizwan Ali Naqvi\inst{4} \and
Radu Timofte\inst{1}
}
\authorrunning{F. A. Dharejo et al.}
\institute{
$^1$University of W\"urzburg, Germany \quad
$^2$Independent Researcher\\
$^3$Mehran UET, Pakistan \quad
$^4$Sejong University, Republic of Korea\\
\vspace{5pt}
\ttfamily Code: \href{https://github.com/sharif-apu/TriFusion-SR}{\color{blue}github.com/sharif-apu/TriFusion-SR}
}

% \author{Anonymized Authors}  %% Added for anonymized MICCAI submission
% \authorrunning{Anonymized Author et al.}
% \institute{Anonymized Affiliations \\
%     \email{email@anonymized.com}}
  
\maketitle              % typeset the header of the contribution
\begin{abstract}

Multimodal medical image fusion facilitates comprehensive diagnosis by aggregating complementary structural and functional information, but its effectiveness is limited by resolution degradation and modality discrepancies. Existing approaches typically perform image fusion and super-resolution (SR) in separate stages, leading to artifacts and degraded perceptual quality. These limitations are further amplified in tri-modal settings that combine anatomical modalities (e.g., MRI, CT) with functional scans (e.g., PET, SPECT) due to pronounced frequency-domain imbalances. We propose TriFusionSR, a wavelet-guided conditional diffusion framework for joint tri-modal fusion and SR. The framework explicitly decomposes multimodal features into frequency bands using the 2D Discrete Wavelet Transform, enabling frequency-aware cross-modal interaction. We further introduce a Rectified Wavelet Features (RWF) strategy for latent coefficient calibration, followed by an Adaptive Spatial-Frequency Fusion (ASFF) module with gated channel–spatial attention to enable structure-driven multimodal refinement. Extensive experiments demonstrate state-of-the-art performance, achieving 4.8–12.4\% PSNR improvement and substantial reductions in RMSE and LPIPS across multiple upsampling scales.
\keywords{Tri-Modal Medical Image Fusion  \and DWT \and SR \and Conditional Diffusion Model.}

% Authors must provide keywords and are not allowed to remove this Keyword section.

\end{abstract}
\section{Introduction}

Medical image fusion combines complementary multi-modal information to improve diagnostic reliability, providing the high-quality, detail-preserving images essential for clinical analysis \cite{zhou2024multimodal}. Although integrating modalities such as PET, CT, MRI, and SPECT \cite{xu2024simultaneous} yields rich anatomical and functional insights, their inherent disparities in resolution, contrast, and noise complicate the fusion process \cite{varghese2024enhancing}. To counter this, fusion techniques have been extensively explored recently \cite{ma2016infrared}. These methods could be broadly categorized into traditional \cite{li2021joint} and deep learning (DL) approaches \cite{ma2022swinfusion}. Traditional methods, such as sparse representation and transform-domain techniques, are efficient but often fail to preserve fine structures \cite{huang2020review,zawish2019brain,azam2022review}. DL methods learn hierarchical cross-modal representations \cite{zhou2023deep}; however, the instability and mode collapse of GANs have motivated a shift toward diffusion models \cite{rombach2022high,ho2020denoising}. These models generate images via progressive denoising, exemplified by Zhao et al.'s DDPM-based hierarchical Bayesian framework \cite{zhao2023ddfm}. Despite this progress, tri-modal medical image fusion remains underexplored, largely confined to traditional techniques like adaptive sparse representation \cite{jie2022tri} and BitonicX filtering \cite{jie2023tri}.

\begin{wrapfigure}{r}{0.4\linewidth}
\vspace{-.9cm}
  \centering
  \includegraphics[width=\linewidth]{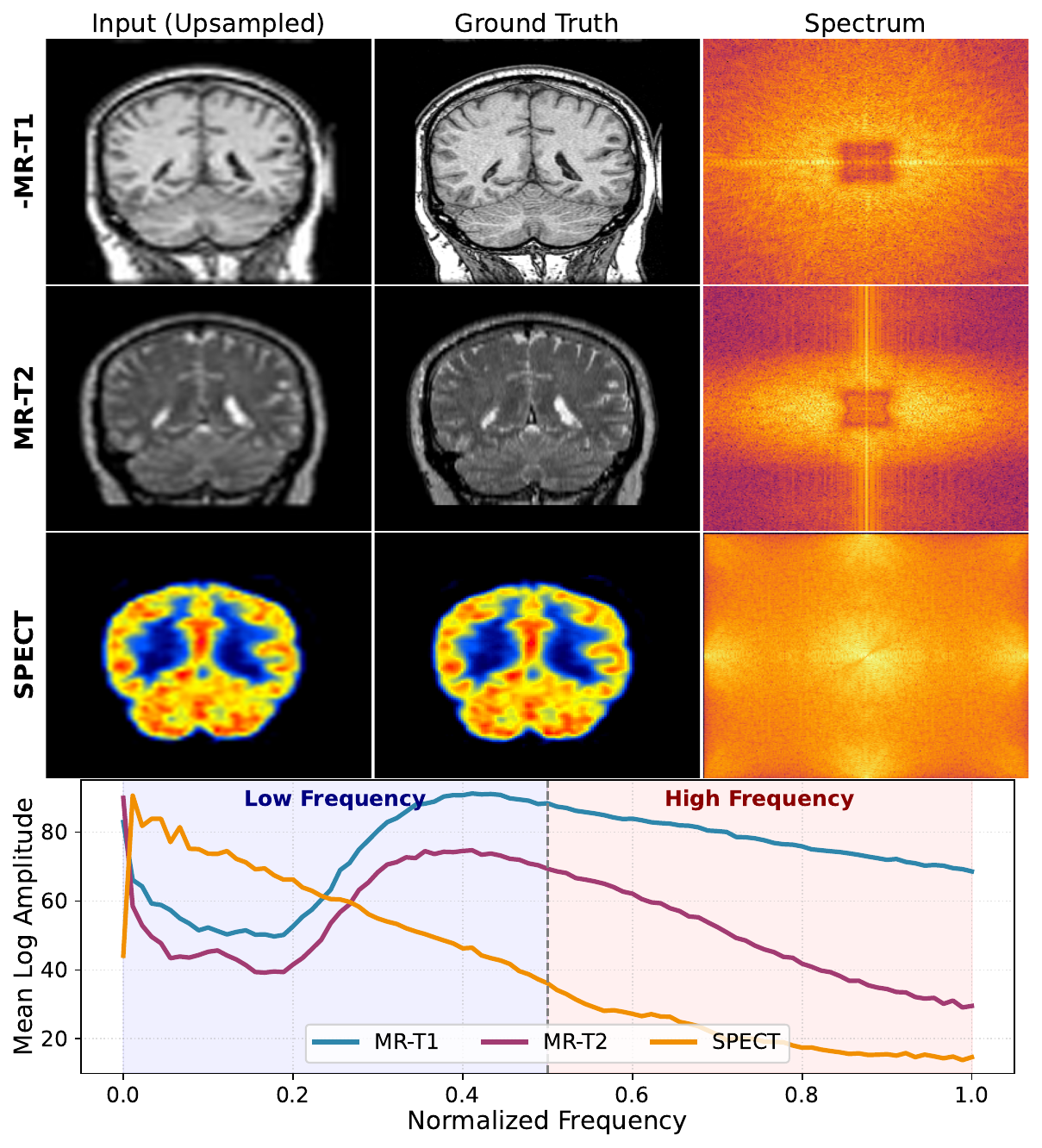}
  \caption{Low–High Frequency Disparity Across MR and SPECT Modalities.}
  \label{fig:intro}
  \vspace{-.8cm}
\end{wrapfigure}
  
Early DL-based SR methods have demonstrated strong capacity for fine-detail recovery with high fidelity~\cite{zeng2018simultaneous,tsiligianni2021interpretable}. In particular, Mao et al. adapted decoupled conditional diffusion frameworks for multi-contrast MRI SR, enabling uncertainty modeling and robust optimization~\cite{ma2020ddcgan,mao2023disc}. However, performing SR and fusion in separate stages can propagate artifacts and degrade final image quality, motivating end-to-end approaches that jointly perform fusion and SR on low-resolution inputs \cite{yin2013simultaneous,ali2025wavehit,xiao2022heterogeneous,li2021joint}. Xiao et al.~\cite{xiao2022heterogeneous} introduced a multi-layer attention-based knowledge distillation framework for infrared–visible fusion and super-resolution, but it is neither designed for medical imaging nor applicable to tri-modal fusion.

Tri-modal medical imaging, particularly combining MRI modalities (e.g.,
MR-T1 and MR-T2) with SPECT/PET further increases complexity due to modality-specific frequency-domain characteristics. For example, as shown in Fig.~\ref{fig:intro}, MRI retains broad high-frequency structural information, while functional signals like SPECT rapidly decay at higher frequencies. This imbalance necessitates frequency-aware handling of low-frequency structural components and high-frequency textures during fusion, which is often overlooked by conventional methods \cite{li2024review}. To address this, we formulate tri-modal fusion and SR by explicitly decomposing features into frequency bands prior to fusion, enabling modality-consistent integration of structural and functional information across diverse imaging combinations. Building on this formulation, we propose TriFusion-SR, a wavelet-guided diffusion model that leverages 2D-DWT for frequency-aware conditioning, enabling simultaneous tri-modal fusion and super-resolution in an end-to-end manner. Our main contributions are summarized as follows:
% \vspace{-.25cm}
\begin{enumerate}
    %\item  We propose leveraging 2D Discrete Wavelet Transform (2D-DWT) into the feature extraction stage to explicitly disentangle structural (low-frequency) and textural (high-frequency) information, providing a principled frequency-domain decomposition that separates signal from noise before cross-modal fusion.
     \item We introduce a novel wavelet diffusion framework that takes advantage of the low-frequency structure and high-frequency details prior to cross-modal fusion. To our knowledge, this is the first end-to-end model to incorporate 2D-DWT into joint tri-modal fusion and SR.

    %\item We introduce a \textbf{Rectified Wavelet Features (RWF)} strategy that projects raw concatenated wavelet coefficients onto a calibrated latent manifold, followed by an ASFF module with joint channel-spatial recalibration and gated refinement, enabling robust adaptive multimodal interaction based on structural relevance rather than noise intensity.
    \item We propose a \textbf{RWF} strategy to calibrate wavelet coefficients in latent space followed by \textbf{ASFF} module with gated channel–spatial attention to enable structure-driven multimodal fusion.
    \item We achieve \textbf{state-of-the-art performance} across three challenging upsampling scales ($2\times$, $4\times$, $8\times$), with \textbf{4.8--12.4\% PSNR improvement}, \textbf{11--33\% RMSE reduction}, and \textbf{52--65\% LPIPS reduction} over existing methods on Joint tri-modal fusion-SR.
\end{enumerate}

        %\item SR
        %\item Joint Fusion-SR
    %\end{itemize}
    %\item P3 -> Problem Formulation (frequency corrupt Fig.1)
    %\begin{itemize}
     %   \item Different Modality impact different frequency band (frequency split)
       % \item Fusion after frequency priors
   % \end{itemize}
    %\item P4 -> Method
    %\begin{itemize}
       % \item frequency split: HF-LF wavelet
       % \item Fusion
       % \begin{itemize}
            %\item RWF (noise-suppressed basis)
            %\item ASFF (channel-spatial gating)
        %\end{itemize}
    %\end{itemize}
    %\item P5 -> Contributions
%\end{itemize}

\section{Methodology}

\subsection{Framework Overview}
\begin{figure}[!htb]
  \centering
  \includegraphics[width=\linewidth]{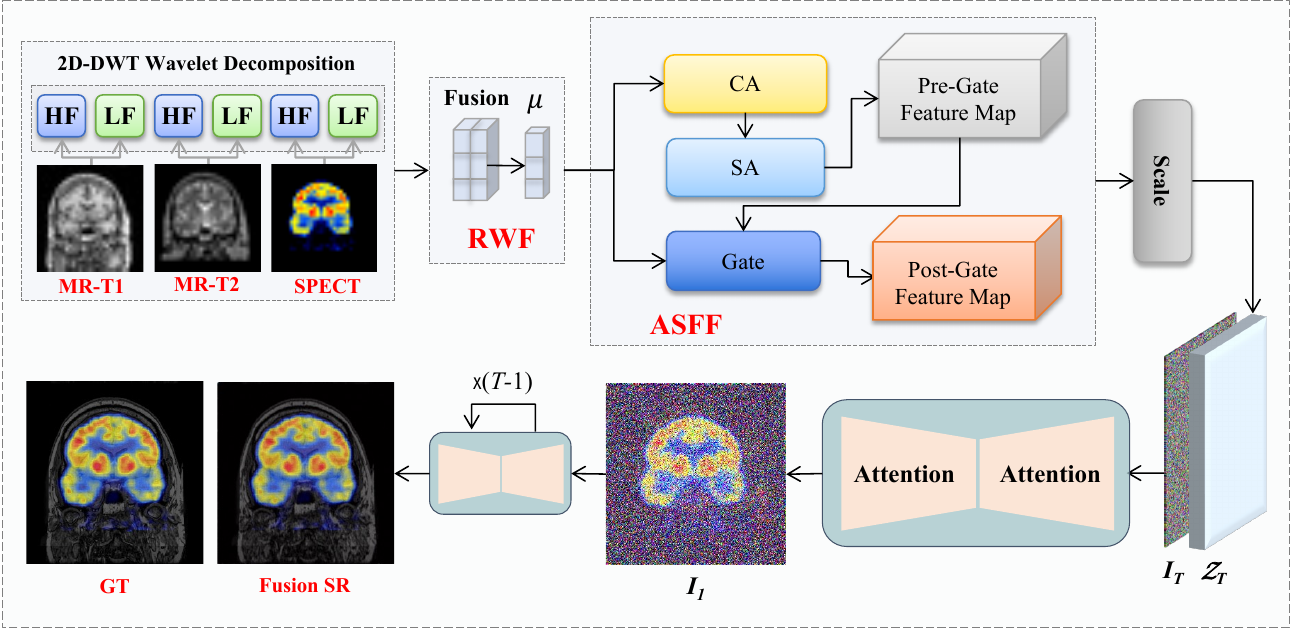}
  \caption{Overview of the proposed TriFusion-SR Framework.}

  \label{fig:network}
  % \vspace{-0.5cm}
\end{figure}

We propose TriFusion-SR, a multimodal image SR framework based on conditional denoising diffusion probabilistic models. The method incorporates a 2DWT to decompose input images into high- and low-frequency components, enabling effective extraction of structural details and texture information. By leveraging frequency-aware fusion within the diffusion process, the model generates high-resolution fused images with enhanced edge preservation and richer fine-grained textures, which are critical for multimodal SR tasks. As illustrated in Fig. \ref{fig:network}, TriFusion-SR achieves accurate tri-modal medical image fusion and SR through a forward and reverse Markov chain diffusion process.

Taking the tri-modal fusion of MR-T1, MR-T2, and SPECT images as an example, the low-resolution inputs from the three modalities are denoted as 
$\mathbf{x} \in \mathbb{R}^{H \times W}$, 
$\mathbf{y} \in \mathbb{R}^{H \times W}$, and 
$\mathbf{s} \in \mathbb{R}^{H \times W}$, respectively. 
The generated high-resolution fused image is denoted by 
$\mathbf{I}_0 \in \mathbb{R}^{3 \times H \times W}$.

The three modal images $\mathbf{x}$, $\mathbf{y}$, and $\mathbf{s}$ are first 
upsampled to the target resolution via bicubic interpolation, producing 
$\tilde{\mathbf{x}}$, $\tilde{\mathbf{y}}$, and $\tilde{\mathbf{s}}$. 
Subsequently, each modality is decomposed using a two-dimensional discrete 
wavelet transform (2D-DWT) to extract low-frequency and high-frequency components:

\begin{equation}
\left\{ \mathbf{x}_L, \mathbf{x}_H \right\} = \mathcal{X}(\tilde{\mathbf{x}}), \quad
\left\{ \mathbf{y}_L, \mathbf{y}_H \right\} = \mathcal{X}(\tilde{\mathbf{y}}), \quad
\left\{ \mathbf{s}_L, \mathbf{s}_H \right\} = \mathcal{X}(\tilde{\mathbf{s}}),
\end{equation}

where $\mathcal{X}(\cdot)$ denotes the 2D-DWT operator.

The decomposed frequency components are then fed into the ASFF block for feature extraction, yielding.

\begin{equation}
\mathbf{z}_t = \varepsilon(\mathbf{x}_L, \mathbf{x}_H, 
\mathbf{y}_L, \mathbf{y}_H, 
\mathbf{s}_L, \mathbf{s}_H).
\end{equation}

The extracted feature $\mathbf{z}_t$ is concatenated along the channel dimension 
with the noisy sample $\mathbf{I}_t \sim \mathcal{N}(0, \mathbf{I})$:

\begin{equation}
\mathbf{h}_t = \text{Concat}(\mathbf{I}_t, \mathbf{z}_t).
\end{equation}

The objective function optimized by Wave-TriFusion is defined as:

\begin{equation}
\mathcal{L}_{\text{(TFS)}} 
= \mathbb{E}_{\mathbf{x}, \mathbf{y}, \mathbf{s}, 
\boldsymbol{\epsilon} \sim \mathcal{N}(0,\mathbf{I}), 
t}
\left[
\left\|
\boldsymbol{\epsilon} -
\boldsymbol{\epsilon}_{\theta}
\left(
\mathbf{z}_t,
\mathbf{I}_t,
\gamma_t
\right)
\right\|_2^2
\right],
\end{equation}

where $\gamma_t$ controls the noise variance in the diffusion process, and $\epsilon_{\theta}$ denotes the U-Net~\cite{ronneberger2015u} parameterized denoising network. The conditioning feature $\mathbf{z}_t$ is produced by the Wavelet-ASFF hybrid module. Our framework adopts the diffusion backbone of TMFS \cite{xu2024simultaneous}, which is based on a U-Net architecture SR3\cite{saharia2022image}. 
%The network comprises a contracting path, an expansive path, and a diffusion head for noise estimation, enhanced with residual connections and self-attention modules. 
We retain the core diffusion architecture and enrich the conditioning branch with wavelet decomposition, RWF and ASFF, to achieve frequency-aware multimodal fusion.
\subsection{Wavelet Decomposition}
%We employ the wavelet transform to perform multi-resolution decomposition, enabling simultaneous capture of spatial localization and frequency information. 
%In contrast to the Fourier transform, which assumes globally uniform frequency behavior, wavelets model spatially varying spectral patterns, making them particularly well-suited for representing structural complexity and textural heterogeneity in medical images.
%By decomposing signals across multiple scales, wavelets preserve fine details while maintaining global structural coherence, making them well-suited for denoising and SR applications. Within our framework, 2D-DWT is embedded into the feature extraction pipeline to explicitly enhance high-frequency structural cues while reducing spatial redundancy, thereby achieving computationally efficient and frequency-aware representation learning.
The discrete wavelet transform (DWT) is widely used to
decompose an image into low-frequency (LF) and high-frequency (HF) sub-bands. In this work, we adopt the Haar wavelet~\cite{klein1923gesammelte}.
Given an input image $\mathbf{I} \in \mathbb{R}^{H \times W \times 3}$,
we apply the DWT operation, denoted as $\mathrm{DWT}(\cdot)$.
This produces one low-frequency sub-band 
$\mathbf{x}_L^{(1)} \in \mathbb{R}^{\frac{H}{2} \times \frac{W}{2} \times 3}$
and three high-frequency sub-bands 
$\{\mathbf{x}_V^{(1)}, \mathbf{x}_H^{(1)}, \mathbf{x}_D^{(1)}\}
\in \mathbb{R}^{\frac{H}{2} \times \frac{W}{2} \times 3}$:
\begin{equation}
\mathbf{x}_L^{(1)}, \mathbf{x}_V^{(1)}, \mathbf{x}_H^{(1)}, \mathbf{x}_D^{(1)}
= \mathrm{DWT}(\mathbf{I}).
\end{equation}
The decomposition can be recursively applied to the low-frequency
sub-band $\mathbf{x}_L^{(1)}$, resulting in:
\begin{equation}
\mathbf{x}_L^{(2)}, \mathbf{x}_V^{(2)}, \mathbf{x}_H^{(2)}, \mathbf{x}_D^{(2)}
= \mathrm{DWT}(\mathbf{x}_L^{(1)}).
\end{equation}

After performing the DWT up to the $J$-th level, we obtain:

\begin{equation}
\{\mathbf{x}_L^{(J)}, \mathbf{x}_V^{(J)}, \mathbf{x}_H^{(J)}, \mathbf{x}_D^{(J)}\}
\in \mathbb{R}^{\frac{H}{2^J} \times \frac{W}{2^J} \times 3}.
\end{equation}

At each level, the low-frequency sub-band is recursively
replaced by its decomposed components~\cite{mallat2002theory}. Therefore,
after $J$ levels, the complete set of sub-bands is:
\begin{equation}
\left\{\mathbf{x}_L^{(J)},\ \mathbf{x}_V^{(J)},\ \mathbf{x}_H^{(J)},\ \mathbf{x}_D^{(J)},\ \mathbf{x}_V^{(J-1)},\ \ldots,\ \mathbf{x}_D^{(1)}\right\}
\end{equation}

Finally, we reshape and concatenate all multi-level sub-bands
to construct the $J$-level wavelet spectrum representation
of the image, denoted as $\mathbf{I}_J^{\mathrm{Recon}}$:

\begin{equation}
\mathbf{I}_J^{\mathrm{Reonc}} = \mathrm{DWT}(\mathbf{I}, J).
\end{equation}

Reversibly, the pixel-domain image $\mathbf{I}$ can be reconstructed through the $J$-level inverse discrete wavelet transform, denoted as $\mathrm{IDWT}$, such that $\mathbf{I} = \mathrm{IDWT}(\mathbf{I}^{\text{Reoc}}_{J}, J)$.

\subsection{Wavelet-Guided Fusion}

% To explicitly preserve structural consistency and fine-grained texture details 
% during multimodal fusion, we introduce a wavelet-guided fusion strategy that 
% operates in the frequency domain prior to diffusion conditioning. To explicitly preserve structural consistency and fine-grained texture details, we introduce a Wavelet-ASFF hybrid framework that operates in the frequency domain prior to diffusion conditioning. Although TMFS~\cite{xu2024simultaneous} unifies fusion and SR, it lacks explicit frequency modeling. We address this by employing 2D-DWT to decompose each modality into structural (low-frequency) and detail (high-frequency) components, enabling spatial–frequency-aware conditioning that separates global anatomy from fine textures.
% To preserve structural consistency and fine-grained texture details during multimodal fusion, we introduce a Wavelet-ASFF hybrid framework that operates in the frequency domain prior to diffusion conditioning. Our approach employs 2D-DWT to decompose each modality into low-frequency (structural) and high-frequency (detail) components, enabling spatial–frequency-aware conditioning that disentangles global anatomy from fine textures.
To explicitly preserve structural consistency and fine-grained texture details during multimodal fusion, we introduce a wavelet-guided fusion strategy that  operates in the frequency domain prior to diffusion conditioning

\subsubsection{Rectified Wavelet Features (RWF)}
Direct concatenation of heterogeneous wavelet sub-bands often introduces \textit{spectral conflict}, where high-frequency noise in functional modalities is conflated with structural details in anatomical ones. To resolve this, we propose a RWF strategy.

Let the set of upsampled low- and high-frequency wavelet components from the three modalities be denoted as $\mathcal{S} = { \mathbf{x}_L, \mathbf{x}_H, \mathbf{y}_L, \mathbf{y}_H, \mathbf{s}L, \mathbf{s}H }$. These components are concatenated to form the raw unified frequency map $\mathbf{F}{\text{raw}} \in \mathbb{R}^{C{in} \times H \times W}$. Rather than processing this noisy union directly, we employ a rectification network $\mathcal{R}(\cdot)$ to project the features into a calibrated latent manifold:

\begin{equation}
\mathbf{F}{\text{rect}} = \mathcal{R}(\mathbf{F}{\text{raw}}).
\end{equation}

Functionally, $\mathcal{R}$ acts as a learnable spectral calibrator. Parameterized by a shared convolutional encoder, it implicitly disentangles stochastic fluctuations (aleatoric noise) from consistent anatomical structures. This yields \textit{Rectified Wavelet Features} $\mathbf{F}_{\text{rect}}$, a noise-suppressed basis that ensures subsequent attention mechanisms are driven by structural correlations rather than inter-modal discrepancies.

\subsubsection{Adaptive Spatial-Frequency Fusion (ASFF)}

% Although TMFS~\cite{xu2024simultaneous} unifies fusion and SR via conditional diffusion but lacks explicit frequency-domain modeling. We address this gap by proposing a Wavelet-ASFF hybrid framework with spatial–frequency-aware conditioning. The 2D-DWT decomposes each modality into structural (low-frequency) and detail (high-frequency) components, separating global anatomy from fine textures. The ASFF module then applies joint channel–spatial attention with gated refinement to enable frequency-aware diffusion conditioning.

% Conceptually, this modification transforms the conditioning variable from a purely semantic embedding into a frequency-structured representation that better aligns with the spectral characteristics of medical images. As a result, both structural fidelity (governed by low-frequency components) and edge sharpness (driven by high-frequency components) are systematically preserved throughout the forward and reverse diffusion processes, leading to improved multimodal consistency and SR accuracy.

% \begin{equation}
% \mathbf{z}_t = \varepsilon(\boldsymbol{\mu}),
% \end{equation}

% where $\varepsilon(\cdot)$ denotes the Hybrid  Wavelet-ASFF feature extractor. 
% The resulting feature $\mathbf{z}_t$ is scaled and used as conditional guidance 
% for the diffusion process.
To refine $\mathbf{F}_{\text{rect}}$, the ASFF module employs a gated channel-spatial attention mechanism. We first generate an attention-enhanced feature map $\mathbf{F}_{\text{sa}} = \mathcal{A}_s(\mathcal{A}_c(\mathbf{F}_{\text{rect}}))$. To dynamically balance structural fidelity with edge enhancement, a gating network $\mathcal{G}$ predicts pixel-wise weights $[w_1, w_2]$ from the concatenated features. The final conditional embedding $\mathbf{z}_t$ is synthesized via gated residual aggregation:

\begin{equation}
\mathbf{z}_t = \underbrace{(w_1 \odot \mathbf{F}_{\text{rect}} + w_2 \odot \mathbf{F}_{\text{sa}})}_{\text{Gated Fusion}} + \underbrace{\gamma \cdot \mathbf{F}_{\text{rect}}}_{\text{Learnable Residual}}
\end{equation}

where $\odot$ denotes element-wise multiplication and $\gamma$ is a learnable parameter. This formulation allows the network to selectively emphasize high-frequency details ($\mathbf{F}_{\text{sa}}$) or preserve original structural information ($\mathbf{F}_{\text{rect}}$) based on local context.

% \subsection{Overall Architecture}

% \subsection{Hybrid Wavelet Fusion Attention}

% \subsection{Fusion SR joint loss function}

\section{Experiments}
\subsection{Experimental Setup}

\paragraph{Dataset and Methods:}
We utilized registered tri-modal image sets from the Harvard Medical School Whole Brain Atlas \cite{summers2003harvard}. The dataset covers five modality combinations: MR-T2/MR-Gad/PET, CT/MR-T2/SPECT, MR-T1/MR-T2/PET, MR-T2/MR-Gad/SPECT, and MR-T1/MR-T2/SPECT. Coronal slices were extracted and normalized to $256 \times 256$ for strict anatomical alignment. Following the TMFS data-split methodology \cite{xu2024simultaneous}, we curated 104 aligned image sets (73 training, 9 validation, 22 testing). For comparison, we evaluated the joint tri-modal framework \textbf{TMFS} \cite{xu2024simultaneous} alongside five state-of-the-art fusion baselines: \textbf{BitonicX} \cite{jie2023tri}, \textbf{CCDFuse} \cite{zhao2023cddfuse}, \textbf{DDFM} \cite{zhao2023ddfm}, \textbf{TGFuse} \cite{rao2023tgfuse}, and \textbf{FlexiD} \cite{xu2025flexid}. Because the latter five are exclusively fusion methods, we coupled them with SR3 \cite{saharia2022image} for super-resolution across $2\times$, $4\times$, and $8\times$ scales.

\paragraph{Implementation Details:}
Implemented in PyTorch, our model was trained across the three aforementioned SR scales: $2\times$ ($128 \rightarrow 256$), $4\times$ ($64 \rightarrow 256$), and $8\times$ ($32 \rightarrow 256$). We optimized the $L_1$ noise prediction loss using AdamW (learning rate $1 \times 10^{-5}$, batch size 4). The diffusion process followed a linear $\beta$-schedule from $10^{-4}$ to $0.02$  over $T=1000$ timesteps. To exploit the RWF strategy, we employed a shared convolutional encoder with 32 initial channels, while the subsequent attention mechanisms utilized a reduction ratio of 16. Training was conducted for 1,000 epochs per scale, with model performance evaluated every 20 epochs with standard metrics. To ensure robustness, we employed a composite weighted score to select the best-performing checkpoint for each scale. Later, we evaluated our test data on the best pretrained weights.

\subsection{Quantitative and Qualitative Results}
% We quantitatively compared our method with state-of-the-art multimodal fusion and SR approaches, including \textbf{BitonicX} \cite{jie2023tri}, \textbf{CCDFuse} \cite{zhao2023cddfuse}, \textbf{DDFM}\cite{zhao2023ddfm}, \textbf{TGFuse}\cite{rao2023tgfuse}, \textbf{FlexiD}\cite{xu2025flexid} and \textbf{TMFS} \cite{xu2024simultaneous}   under $\times 2$, $\times 4$, and $\times 8$ scaling factors.
Table~\ref{tab:quantitative_miccai_final} summarizes quantitative results on the Harvard dataset. Performance was evaluated using PSNR, SSIM, RMSE, and LPIPS to assess pixel-level accuracy, structural fidelity, and perceptual quality. Across all scales and evaluation metrics, our method consistently achieves the best performance. As shown at $\times 2$, compared with the strongest competing method (TMFS), our approach improves PSNR by $12.35\%$ and SSIM by $1.61\%$, while reducing RMSE by $32.51\%$ and LPIPS by $65.46\%$. The notable decrease in LPIPS and RMSE reflects enhanced structural fidelity and perceptual consistency, supporting the robustness of the proposed framework. 

\begin{table}[t]
\centering
\caption{Quantitative comparison on the test set sorted by scaling factors. \textbf{Bold} indicates the best performance, and \underline{underlining} indicates the second-best results.}
\label{tab:quantitative_miccai_final}
% Using \small instead of resizebox for better font consistency
\small 
\setlength{\tabcolsep}{3.5pt} % Adjust column spacing to fit width
\scalebox{0.85}{\begin{tabular}{c|l|cccc}
\toprule
\textbf{Scale} & \textbf{Method} & \textbf{PSNR} $\uparrow$ & \textbf{SSIM} $\uparrow$ & \textbf{RMSE} $\downarrow$ & \textbf{LPIPS} $\downarrow$ \\
\midrule
\multirow{6}{*}{\rotatebox{90}{\textbf{Scale $2\times$}}} 
& BitonicX \cite{jie2023tri} & $12.69 \pm 0.50$ & $0.6411 \pm 0.0485$ & $0.2323 \pm 0.0130$ & $0.1817 \pm 0.0226$ \\
& CCDFuse \cite{zhao2023cddfuse}       & $11.46 \pm 0.59$ & $0.6520 \pm 0.0493$ & $0.2680 \pm 0.0177$ & $0.2152 \pm 0.0319$ \\
& TGFuse \cite{rao2023tgfuse}   & $12.61 \pm 0.56$ & $0.2097 \pm 0.0423$ & $0.2345 \pm 0.0140$ & $0.3081 \pm 0.0348$ \\
& DDFM \cite{zhao2023ddfm}       & $20.40 \pm 1.48$ & $0.8315 \pm 0.1826$ & $0.0970 \pm 0.0186$ & \underline{$0.1224 \pm 0.0193$} \\
& FlexiD \cite{xu2025flexid}   & $17.54 \pm 0.69$ & $0.8024 \pm 0.0179$ & $0.1331 \pm 0.0109$ & $0.1246 \pm 0.0228$ \\
& TMFS   \cite{xu2024simultaneous}        & \underline{$27.93 \pm 0.77$} & \underline{$0.8431 \pm 0.0331$} & \underline{$0.0403 \pm 0.0033$} & $0.1248 \pm 0.0238$ \\
\cmidrule{2-6}
\rowcolor{oursgray}
& \textbf{Ours}              & \textbf{31.38} $\pm$ \textbf{1.13} & \textbf{0.8567} $\pm$ \textbf{0.0825} & \textbf{0.0272} $\pm$ \textbf{0.0037} & \textbf{0.0431} $\pm$ \textbf{0.0134} \\
\midrule
\midrule
\multirow{6}{*}{\rotatebox{90}{\textbf{Scale $4\times$}}} 
& BitonicX \cite{jie2023tri} & $12.85 \pm 0.48$ & $0.6124 \pm 0.0540$ & $0.2281 \pm 0.0123$ & $0.2664 \pm 0.0288$ \\
& CCDFuse \cite{zhao2023cddfuse}       & $11.49 \pm 0.61$ & $0.6258 \pm 0.0520$ & $0.2669 \pm 0.0184$ & $0.3362 \pm 0.0374$ \\
& TGFuse \cite{rao2023tgfuse}   & $12.55 \pm 0.55$ & $0.2055 \pm 0.0370$ & $0.2361 \pm 0.0139$ & $0.3702 \pm 0.0445$ \\
& DDFM \cite{zhao2023ddfm}       & $19.27 \pm 1.43$ & \underline{$0.7793 \pm 0.0264$} & $0.1104 \pm 0.0206$ & $0.2521 \pm 0.0313$ \\
& FlexiD \cite{xu2025flexid}   & $16.92 \pm 0.60$ & $0.7275 \pm 0.0294$ & $0.1430 \pm 0.0099$ & $0.2598 \pm 0.0304$ \\
& TMFS  \cite{xu2024simultaneous}         & \underline{$26.22 \pm 1.47$} & $0.6022 \pm 0.0771$ & \underline{$0.0495 \pm 0.0075$} & \underline{$0.1736 \pm 0.0291$} \\
\cmidrule{2-6}
\rowcolor{oursgray}
& \textbf{Ours}              & \textbf{27.62} $\pm$ \textbf{1.14} & \textbf{0.7864} $\pm$ \textbf{0.1004} & \textbf{0.0420} $\pm$ \textbf{0.0056} & \textbf{0.0839} $\pm$ \textbf{0.0305} \\
\midrule
\midrule
\multirow{6}{*}{\rotatebox{90}{\textbf{Scale $8\times$}}} 
& BitonicX \cite{jie2023tri} & $13.08 \pm 0.47$ & $0.5355 \pm 0.0731$ & $0.2220 \pm 0.0117$ & $0.4066 \pm 0.0430$ \\
& CCDFuse \cite{zhao2023cddfuse}       & $11.60 \pm 0.61$ & $0.5480 \pm 0.0657$ & $0.2638 \pm 0.0184$ & $0.4888 \pm 0.0588$ \\
& TGFuse \cite{rao2023tgfuse}   & $12.63 \pm 0.54$ & $0.1825 \pm 0.0297$ & $0.2342 \pm 0.0136$ & $0.4695 \pm 0.0423$ \\
& DDFM \cite{zhao2023ddfm}       & $18.38 \pm 1.42$ & \underline{$0.6149 \pm 0.1415$} & $0.1222 \pm 0.0230$ & $0.3974 \pm 0.0472$ \\
& FlexiD \cite{xu2025flexid}   & $16.36 \pm 0.49$ & $0.6108 \pm 0.0561$ & $0.1523 \pm 0.0086$ & $0.4052 \pm 0.0449$ \\
& TMFS  \cite{xu2024simultaneous}         & \underline{$24.84 \pm 1.12$} & $0.5978 \pm 0.0999$ & \underline{$0.0577 \pm 0.0073$} & \underline{$0.1943 \pm 0.0339$} \\
\cmidrule{2-6}
\rowcolor{oursgray}
& \textbf{Ours}              & \textbf{26.04} $\pm$ \textbf{1.85} & \textbf{0.6287} $\pm$ \textbf{0.1333} & \textbf{0.0511} $\pm$ \textbf{0.0118} & \textbf{0.0835} $\pm$ \textbf{0.0264} \\
\bottomrule
\end{tabular}}%
\end{table}

\begin{figure}[!htb]
  \centering
  \includegraphics[width=\linewidth]{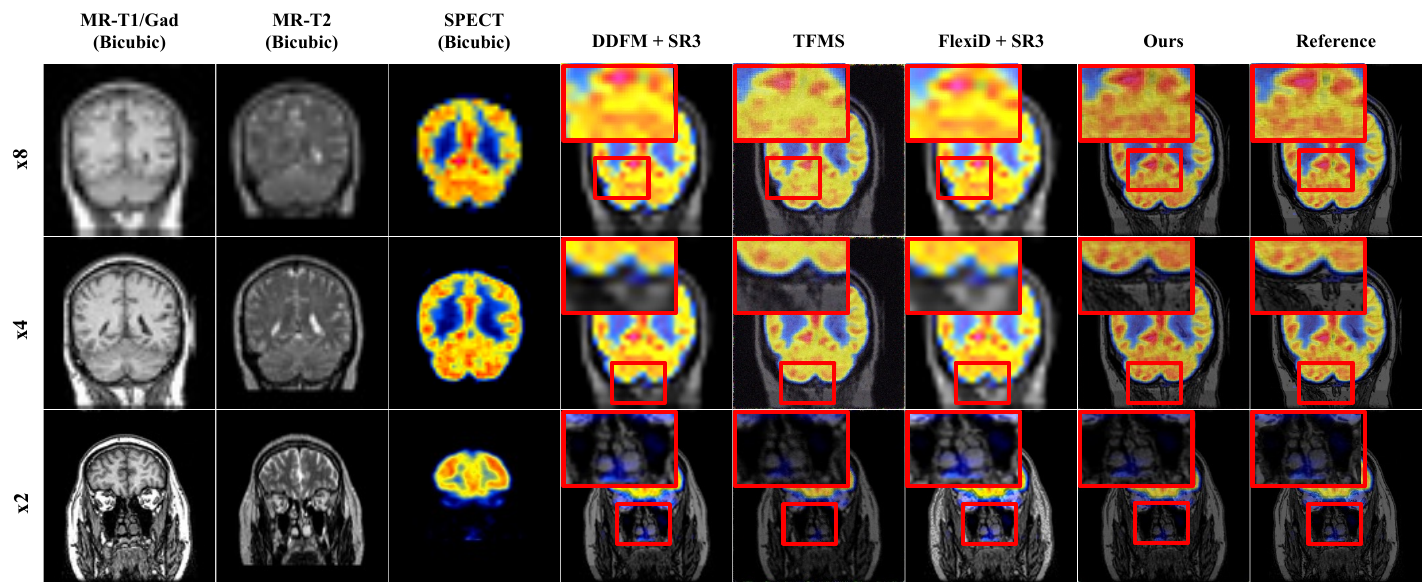}
  \caption{TriFusion-SR qualitative results at $\times 2$, $\times 4$, and $\times 8$ on the Harvard dataset.}

  \label{fig3}
  % \vspace{-0.5cm}
\end{figure}

The qualitative results in Fig. \ref{fig3} further validate these findings. As the scaling factor increases, DDFM\cite{zhao2023ddfm} , TFMS \cite{xu2024simultaneous}, and Flexid \cite{xu2025flexid} exhibit noticeable texture blurring and structural degradation. While TriFusion-SR maintains general structural consistency, produces sharper boundaries, more accurate color representation, and richer textures that more closely match the ground truth, particularly under high upscaling settings.

\subsection{Ablation Study}
% This section evaluates the effectiveness of the Wavelet decomposition, ASFF module, and the combined Wavelet--ASFF--RWF hybrid framework for SR.

% \textbf{Wavelet Only (w/o ASFF Block):} 
% Compared with the baseline, using Wavelet decomposition alone significantly improves PSNR to 27.28, corresponding to a 14.48\% increase, and raises SSIM by 16.82\%. This setting removes the ASFF module to assess its contribution to feature extraction and multimodal information fusion.

%  \textbf{Wavelet + ASFF:} 
% Introducing the ASFF module further enhances perceptual and structural consistency. Specifically, LPIPS decreases from 0.130 to 0.098 (a 24.62\% reduction), while SSIM improves from 0.778 to 0.835, demonstrating the effectiveness of adaptive spatial–frequency recalibration.

% \textbf{Wavelet + RWF + ASFF (Proposed):} 
% Integrating RWF with Wavelet and ASFF achieves the best overall performance (PSNR: 27.62, SSIM: 0.786, RMSE: 0.0420, LPIPS: 0.084). Although SSIM slightly decreases compared with the Wavelet+ASFF configuration, PSNR and LPIPS further improve while RMSE remains same. 

% The effect of the Wavelet and ASFF modules on visual representations is shown in Fig.~\ref{Ablation}. The ablation study in Table~2 highlights the complementary roles of the Wavelet, ASFE, and RWF modules, each contributing to measurable performance gains in tri-modal fusion SR. The combination of these components enhances both pixel-level reconstruction accuracy and perceptual quality, providing a robust solution for complex medical image SR tasks.
\begin{figure}[!htb]
  \centering
  \includegraphics[width=.9\linewidth]{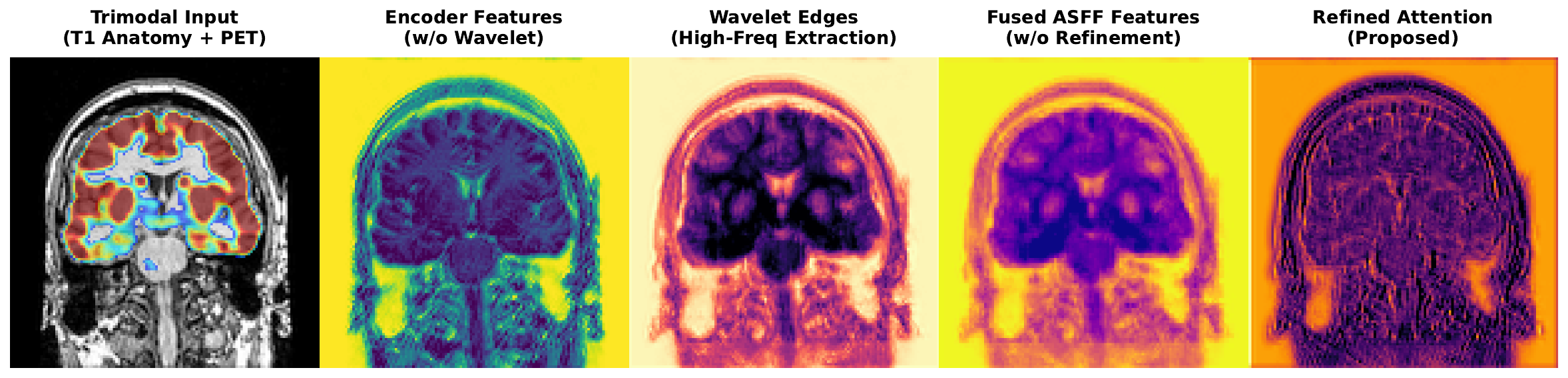}
  \caption{Impact of Wavelet and ASFF Modules on Feature Representation.}

  \label{Ablation}

\end{figure}

% \begin{wrapfigure}{r}{0.7\linewidth}
% \vspace{-.9cm}
%   \centering
%   \includegraphics[width=\linewidth]{images/attention_ablation_eval.pdf}
%   \caption{Impact of Wavelet and ASFF Modules on Feature Representation.}
%   \label{Ablation}
%   \vspace{-.8cm}
% \end{wrapfigure}

To assess the contribution of individual components, we evaluated the wavelet decomposition, ASFF module, and the proposed RWF strategy, as detailed in Table~2. Compared to the baseline, utilizing wavelet decomposition alone yielded a significant performance boost, increasing PSNR to 27.28 (a 14.48\% gain) and raising SSIM by 16.82\%. The addition of the ASFF module further enhanced structural and perceptual consistency, reducing LPIPS by 24.62\% (from 0.130 to 0.098) and improving SSIM to 0.835. The full framework comprising Wavelet with  RWF and ASFF achieved the best overall performance (PSNR: 27.62, LPIPS: 0.084, RMSE: 0.0420). Notably, the proposed RWF resulted in a slight decrease in SSIM (0.786) compared to the Wavelet+ASFF configuration; however, we strategically trade off structural metrics for perceptual fidelity. This balance optimizes pixel-level accuracy and visual detail, both essential for reliable medical diagnosis. As illustrated in Fig.~\ref{Ablation}, these modules play complementary roles, providing a robust solution for complex tri-modal medical image SR.

\begin{table}[!htb]
\centering
\caption{Ablation study on different model configurations. \textbf{Bold} indicates the best performance.}
\label{tab:ablation}
\small 
\setlength{\tabcolsep}{3.5pt}
\scalebox{0.7}{\begin{tabular}{lccc|cccc}
\toprule
\multirow{2}{*}{\textbf{Method}} & \multicolumn{3}{c|}{\textbf{Components}} & \multicolumn{4}{c}{\textbf{Metrics}} \\
\cmidrule(lr){2-4} \cmidrule(lr){5-8}
 & \textbf{Wavelet} & \textbf{RWF} & \textbf{ASFF} & \textbf{PSNR} $\uparrow$ & \textbf{SSIM} $\uparrow$ & \textbf{RMSE} $\downarrow$ & \textbf{LPIPS} $\downarrow$ \\
\midrule
BASE & \xmark & \xmark & \xmark & $23.83 \pm 1.79$ & $0.666 \pm 0.149$ & $0.0658 \pm 0.0144$ & $0.176 \pm 0.049$ \\
Wavelet & \cmark & \xmark & \xmark & $27.28 \pm 0.87$ & $0.778 \pm 0.088$ & $0.0435 \pm 0.0043$ & $0.130 \pm 0.030$ \\
Wavelet-ASFF & \cmark & \xmark & \cmark & $27.59 \pm 0.91$ & $0.835 \pm 0.026$ & $0.0420 \pm 0.0043$ & $0.098 \pm 0.028$ \\
\cmidrule{1-8}
\rowcolor{oursgray}
\textbf{Proposed} & \cmark & \cmark & \cmark & $\mathbf{27.62} \pm \mathbf{1.14}$ & $\mathbf{0.786} \pm \mathbf{0.100}$ & $\mathbf{0.0420} \pm \mathbf{0.0056}$ & $\mathbf{0.084} \pm \mathbf{0.031}$ \\
\bottomrule
\end{tabular}}
\end{table}
\section{Conclusion}
We proposed \textbf{TriFusion-SR}, a wavelet-guided conditional diffusion framework for tri-modal medical image fusion and SR. By integrating 2D-DWT into feature extraction, the framework disentangles low-frequency structural and high-frequency textural components prior to cross-modal interaction. The proposed RWF strategy projects concatenated wavelet coefficients onto a calibrated latent manifold, while the ASFF module jointly recalibrates channel–spatial responses with gated refinement, collectively suppressing noise-driven responses during diffusion conditioning. Quantitative, visual, and ablation results confirm the synergistic contribution of each component and the superiority of TriFusion-SR over existing methods. In future work, we aim to incorporate foundation models to provide stronger semantic priors for diffusion-guided learning and to improve generalization across diverse modalities and clinical scenarios.
% In this work, we presented a wavelet-guided conditional diffusion framework, TriFusion-SR, for tri-modal medical image fusion and SR. The proposed method integrates 2D-DWT into the feature extraction stage to explicitly disentangle low-frequency structural components from high-frequency textural details, enabling principled frequency-domain decomposition prior to cross-modal interaction. To further enhance SR quality, we introduced a Rectified Wavelet Features (RWF) strategy that projects concatenated wavelet coefficients onto a calibrated latent manifold, followed by an ASFF module that performs joint channel–spatial recalibration with gated refinement. This hybrid design promotes structurally relevant multimodal fusion while suppressing noise-driven responses during diffusion conditioning. Both quantitative and visual comparisons demonstrate that TriFusion-SR outperforms existing methods. Ablation results highlight the synergistic impact of wavelet-based decomposition, RWF , and ASFF components. 

% In future work, we aim to incorporate foundation models to provide stronger semantic priors for diffusion guidance and improve generalization across diverse medical imaging modalities and clinical scenarios.

%\begin{credits}
%\subsubsection{\ackname} This work was supported by The
%Alexander von Humboldt Foundation.

%\subsubsection{\discintname}
%The authors report no conflicts of interest relevant to this study.
%\end{credits}

\bibliographystyle{splncs04}





% LNCS Homepage, \url{http://www.springer.com/lncs}, last accessed 2023/10/25
% \end{thebibliography}
\end{document}